\documentclass[runningheads]{llncs}
\usepackage{graphicx}
\usepackage{comment}
\usepackage{amsmath,amssymb} 
\usepackage{color}

\usepackage[width=122mm,left=12mm,paperwidth=146mm,height=193mm,top=12mm,paperheight=217mm]{geometry}

\begin{document}
\pagestyle{headings}
\mainmatter

\title{Learning Disentangled Feature Representation for Hybrid-distorted Image Restoration} 



\titlerunning{Learning Disentangled Feature Representation for HD-IR}
%
\author{Xin Li \and
Xin Jin \and Jianxin Lin \and Tao Yu \and Sen Liu  \and Yaojun Wu \and Wei Zhou \and Zhibo Chen \thanks{Corresponding Author}}
\authorrunning{X. Li et al.}
%
\institute{CAS Key Laboratory of Technology in Geo-spatial Information Processing and Application System,\\
	University of Science and Technology of China, Hefei 230027, China\\
\email{\{lixin666,jinxustc,linjx\}@mail.ustc.edu.cn},
\email{elsen@iat.ustc.edu.cn}, \email{\{yaojunwu,yutao666,weichou\}@mail.ustc.edu.cn}, \email{chenzhibo@ustc.edu.cn} \\
}
\maketitle

\begin{abstract}
Hybrid-distorted image restoration (HD-IR) is dedicated to restore real distorted image that is degraded by multiple distortions.
Existing HD-IR approaches usually ignore the inherent interference among hybrid distortions which compromises the restoration performance. To decompose such interference, we introduce the concept of Disentangled Feature Learning to achieve the feature-level divide-and-conquer of hybrid distortions. Specifically, we propose the feature disentanglement module (FDM) to distribute feature representations of different distortions into different channels by revising gain-control-based normalization.  We also propose a feature aggregation module (FAM) with channel-wise attention to adaptively filter out the distortion representations and aggregate useful content information from different channels for the construction of raw image. The effectiveness of the proposed scheme is verified by visualizing the correlation matrix of features and channel responses of different distortions. Extensive experimental results also prove superior performance of our approach compared with the latest HD-IR schemes. 
\end{abstract}

\section{Introduction}
 \begin{figure}[h]
 	\centering
 	\includegraphics[width=\linewidth]{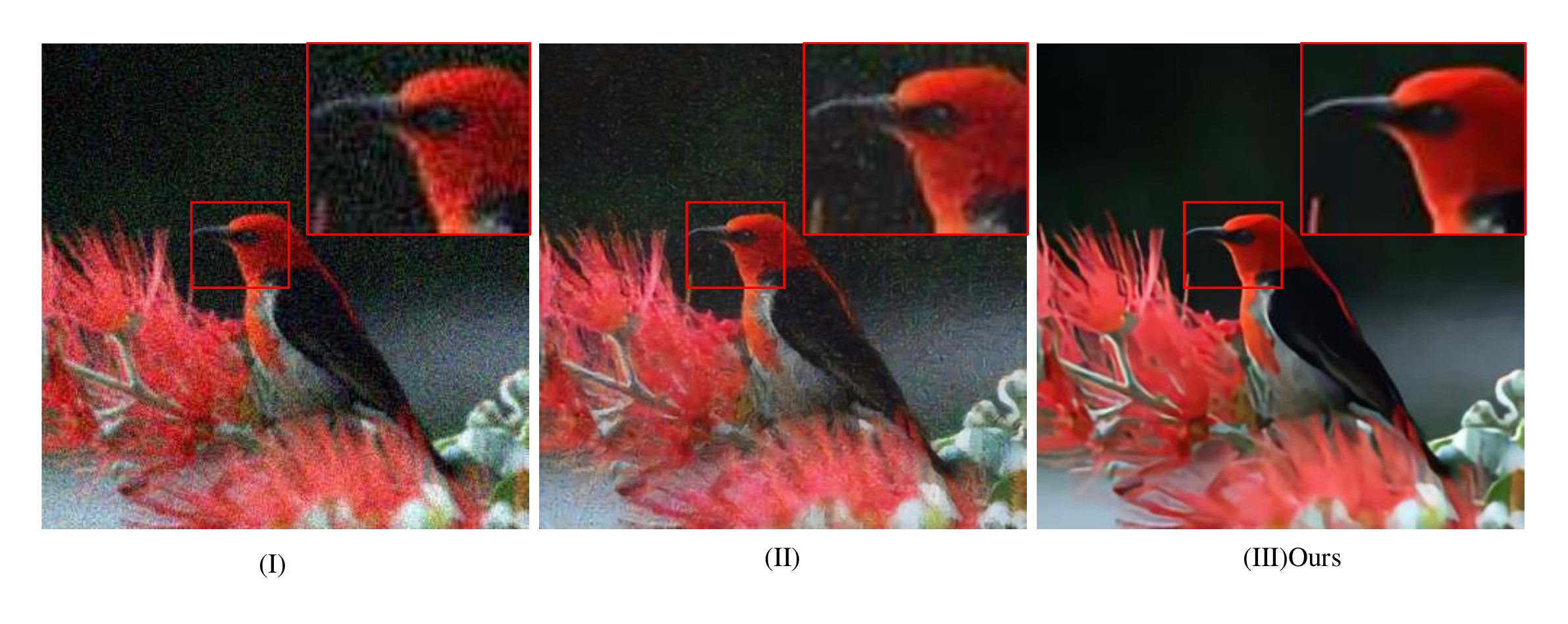}
 	\caption{Examples of hybrid-distorted image restoration. (I) Hybrid-distorted image including noise, blur, and jpeg artifacts. (II) Processed with the cascading single distortion restoration networks including dejpeg, denoise and deblurring. (III)Processed with our FDR-Net.}
 	\label{fig:distortion_example}
 \end{figure}

 Nowadays, Image restoration techniques have been applied in various fields, including streaming media, photo processing, video surveillance, and cloud storage, etc. In the process of image acquisition and transmission, raw images are usually contaminated with various distortions due to capturing devices, high ratio compression, transmission, post-processing, etc.
 Previous image restoration methods focusing on single distortion have been extensively studied \cite{dong2015image,kim2016accurate,yu2018generative,nazeri2019edgeconnect,kupyn2018deblurgan,fergus2006removing,shi2019low,lehtinen2018noise2noise,li2018non,yang2018towards} and achieved satisfactory performance on the field of super resolution \cite{lim2017enhanced,haris2018deep,li2019feedback} , deblurring \cite{tao2018scale,lu2019unsupervised,yuan2019blind}, denoising \cite{dabov2007image,zhang2017beyond,zhang2018ffdnet}, deraining \cite{jin2019unsupervised,jin2018decomposed,yang2017deep}, dehazing \cite{zhao2019dd,berman2016non,zhang2017joint} and so on. However, these works are usually designed for solving one specific distortion, which makes them difficult to be applied to real world applications as shown in Fig. \ref{fig:distortion_example}.
 
 Real world images are typically affected by multiple distortions simultaneously. 
 In addition, different distortions might be interfered with each other,  which makes it difficult to restore images. More details about interference between different distortions can be seen in the  supplementary material. Recently, there have been proposed some pioneering works for hybrid distortion. For example, Yu $et$ $al.$ \cite{yu2018crafting}  pre-train several light-weight CNNs as tools for different distortions. Then they utilize the Reinforcement Learning (RL) agent to learn to choose the best tool-chain for unknown hybrid distortions. Then Suganuma $et$ $al.$ \cite{suganuma2019attention} propose to use the attention mechanism to implement the adaptive selection of different operations for hybrid distortions. However, these methods are designed regardless of the interference between hybrid distortions.

\begin{figure}[htp]
	\centering
	\includegraphics[width=\linewidth]{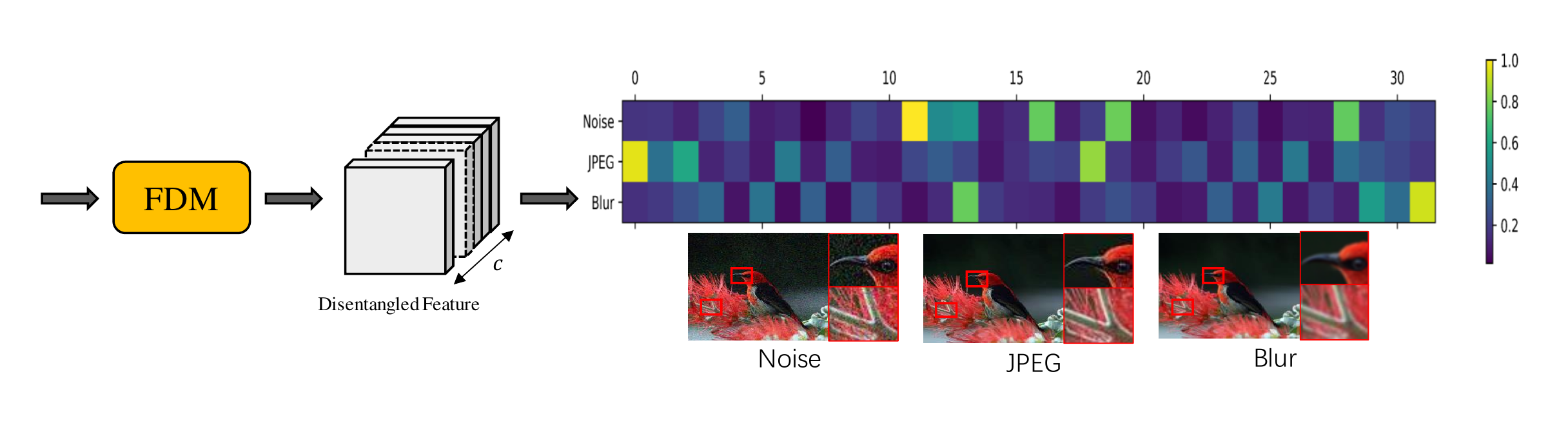}
	\caption{Illustration of feature disentangle module (FDM) works. The FDM disentangles the input feature representation which characterizes the hybrid-distorted image into disentangled features. The responses of different distortions are distributed across different channels as shown in brighter region. The visualizations of distortions are also displayed.}
	\label{fig:shiyi}
\end{figure}

Previous literature \cite{Bianco2020DisentanglingID} suggests that deep feature representations could be employed to efficiently characterize various image distortions. In other words, the feature representation extracted from hybrid distortions images might be disentangled to characterize different distortions respectively. Based on the above, we propose to implement the feature-level divide-and-conquer of the hybrid distortions by learning disentangled feature representations. On a separate note, Schwartz $et$ $al.$ \cite{schwartz2001natural}  point out that a series of filters and gain-control-based normalization could achieve the decomposition of different filter responses. And the convolution layer of CNN is also composed of a series of basic filters/kernels. Inspired by this, we expand this theory and design a feature disentanglement module (FDM) to implement the channel-wise feature decorrelation.
By such feature decorrelation, the feature representations of different distortions could be distributed across different channels respectively as shown in Fig. \ref{fig:shiyi}. 

Different from high-level tasks such as classification \cite{liu2018bi,he2016deep}, person re-identi-
fication \cite{li2018unsupervised,chen2019abd,jin2020style}, the processed feature representations are not clustered into several classes. It is crucial for low-level image restoration to aggregate the useful content information from processed feature representations to reconstruct the image. Therefore, we design an adaptive feature aggregation module (FAM) based on channel-wise attention mechanism and inverse transform of gain-control-based normalization \cite{schwartz2001natural}. 

Extensive experiments are conducted on hybrid distortion datasets and our model achieves the state-of-the-art results. Furthermore, we give a reasonable interpretation for our framework through visualizing the correlation matrix of features and channel response of different distortions. Finally, we verify the robustness of our network on single distortion tasks such as image deblurring and image deraining. 

Our main contributions can be summarized as follows:
\begin{itemize}
	\item We implement the feature-level divide-and-conquer of hybrid-distortion by feature disentanglement. And we design the  Feature disentangled module (FDM) by revising gain-control-based normalization to distribute feature representations of different distortions into different channels.  
	\item We propose a Feature Disentanglement Network by incorporating FDM and FAM for hybrid-distorted image restoration named FDR-Net.
	\item Extensive experiments demonstrate that our FDR-Net has achieved the state-of-the-art results on hybrid-distorted image restoration. We also verify the efficiency of our modules through ablation study and visualization.
\end{itemize}

\section{Related Work}
In this section, we briefly summarize the researches on single-distorted image restoration and hybrid-distorted image restoration. 
\subsection{Image Restoration on Single Distortion}
Deep learning has promoted the development of computer vision tasks, especially on image restoration. As the pioneers in image restoration, some early works on special distortion removal have been proposed. Dong $et$ $al.$ first employ a deep convolutional neural network (CNN) in image super-resolution in \cite{dong2015image}. Zhang $et$ $al.$ propose Dncnn \cite{zhang2017beyond} to handle noise and JPEG compression distortion. And Sun $et$ $al.$ utilize CNN to estimate the motion blur kernels for image deblur in \cite{sun2015learning}. With the advance of CNN and the improvement of computing resources, the depth and the number of parameters of the neural network have been increased significantly which facilitates a series of excellent works on special tasks such as deblurring \cite{tao2018scale,lu2019unsupervised,yuan2019blind}, super-resolution \cite{lim2017enhanced,haris2018deep,li2019feedback}, deraining \cite{jin2019unsupervised,jin2018decomposed,yang2017deep} and so on to improve the image quality.

To further improve the subjective quality of degraded images, the generative adversarial network has been utilized to restore image restoration. SRGAN \cite{ledig2017photo}, DeblurGAN \cite{kupyn2018deblurgan}, ID-CGAN \cite{zhang2019image} have employed GAN in the tasks of super-resolution, deblurring, and deraining respectively to generate more realistic images. Yuan $et$ $al.$ \cite{yuan2018unsupervised} use Cycle-in-Cycle Generative Adversarial Networks to implement unsupervised image super-resolution. Ulyanov $et$ $al.$ \cite{ulyanov2018deep} prove that the structure of generator network is sufficient to capture low-level image statistics prior and then utilize it to implement unsupervised image restoration.


\subsection{Image Restoration on Hybrid Distortion}
Recently, with the wide applications of image restoration, the special image restoration on single distortion cannot meet the need of real world application. Then, some works on hybrid distortion restoration have been proposed. Among them, RL-Restore \cite{yu2018crafting} trains a policy to select the appropriate tools for single distortion from the pre-trained toolbox to restore the hybrid-distorted image. Then Suganuma $et$ $al.$ \cite{suganuma2019attention} propose a simple framework which can select the proper operation with attention mechanism. However, these works don't consider the interference between different distortions. In this paper, we propose the FDR-Net for hybrid distortions image restoration by reducing the interference with feature disentanglement, which achieves more superior performance by disentangling the feature representation for hybird-distorted image restoration.

\section{Approach}
In this section, we start with introducing the gain control based signal decomposition as primary knowledge for feature disentanglement module (FDM), and then elaborate the components of our FDR-Net and its overall architecture.

\subsection{Primary Knowledge}
 To derive the models of sensory processing, gain-control-based normalization has been proposed to implement the nonlinear decomposition of natural signals \cite{schwartz2001natural}. Given a set of signals $\vec{X}$, the signal decomposition first uses a set of filters $\left\langle {{f}_{1}},{{f}_{2}},...,{{f}_{n}}   \right\rangle $  to extract different representations of signals  $\left\langle {{L}_{1}},{{L}_{2}},...{{L}_{n}}\right\rangle$ as 
\label{primary}
\begin{equation}
{{L}_{i}}={{f}_{i}}(\vec{X}), i=1,2,...,n, 
\label{filter}
\end{equation}
and then uses gain-control-based normalizaiton as Eq. \ref{gdn} to further eliminate the dependency between $\left\langle {{L}_{1}},{{L}_{2}},...{{L}_{n}} \right\rangle $.
\begin{equation}
{{R}_{i}}=\frac{L_{i}^{2}}{\sum\nolimits_{j}{{{w}_{ji}}L_{j}^{2}}\text{+}{{\sigma }^{\text{2}}}},
\label{gdn}
\end{equation}
where independent response ${{R}_{i}}$ can be generated based on the suitable weights ${{w}_{ji}}$ and offsets ${{\sigma }^{\text{2}}}$ with corresponding inputs $L_{i}$. In this way, the input signals  $\vec{X}$ can be decomposed into  ${{R}_{1}},{{R}_{2}},...{{R}_{n}}$ according to their statistical property.

\subsection{Feature Disentanglement}
Previous literature \cite{Bianco2020DisentanglingID} has proved that different distortions have different deep feature representation and could be disentangled at feature representation. Based on this analysis, we design the FDM by expanding the signal decomposition into channel-wise feature disentanglement to reduce the interference between hybrid distortions. As shown in \cite{schwartz2001natural}, the combination of diverse linear filters and divisive normalization
has the capability to decompose the responses of filters. And the feature representation of CNN is also composed of responses from a series of basic filters/kernels. Based on such observation, we implement the adaptive channel-wise feature decomposition by applying such algorithm in learning based framework.
\begin{figure}[h]
	\centering
	\includegraphics[width=\linewidth]{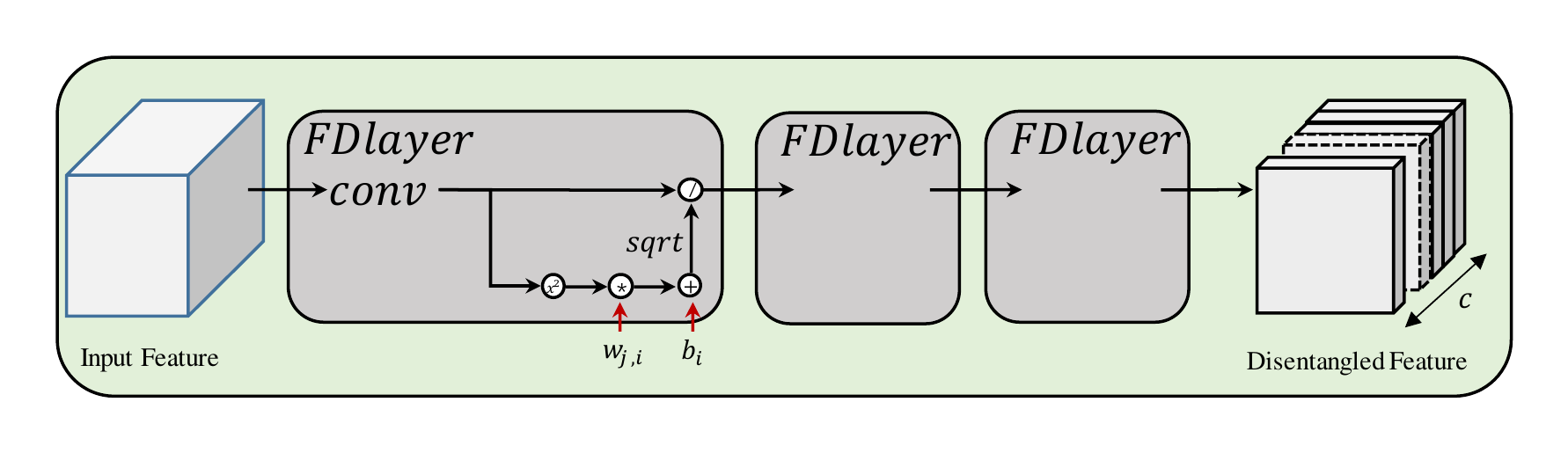}
	\caption{Feature Disentanglement Module.}
	\label{fig:FGmodel}
\end{figure} 
Specifically, we use the convolution layer (${C}_{in}\times{C}_{out}\times{k}\times{k}$) in neural network to replace traditional filters $\left\langle {{f}_{1}},{{f}_{2}},...,{{f}_{n}} \right\rangle $ in signal decomposition as section \ref{primary}. Here, the number of input channel ${C}_{in}$ represents the channel dimension of input feature ${{\vec{F}}_{in}}$, the number of output channel ${C}_{out}$ represents the channel dimension of output feature ${{\vec{F}}_{out}}$, and $k$ is the kernel size of filters. In this way, the extraction results ${{S}_{1}},{{S}_{2}},...{{S}_{{C}_{out}}}$ of convolution layer will be distributed in different channels as:
\begin{equation}
\left\langle {{S}_{1}},{{S}_{2}},...,{{S}_{{{c}_{out}}}} \right\rangle =conv({{\vec{F}}_{in}}),
\end{equation} 
where ${S}_{i}$ represents the $i$th channel in output feature and $conv$ represents the convolution layer.

To introduce the gain-control-based normalization as Eq. \ref{gdn} into CNN, we modified the Eq. \ref{gdn} as Eq. \ref{ftrans}.
\begin{equation}
{{D}_{i}}=\frac{{{S}_{i}}}{\sqrt{\sum\nolimits_{j}{{{w}_{ji}}S_{j}^{2}+{{b}_{i}}}}},
\label{ftrans}
\end{equation}

where ${w}_{ji}$ and ${b}_{i}$ can be learned by gradient descent. ${{S}_{i}}$ and ${{D}_{i}}$ represent the $i$th channel components of features before and after gain control.
In formula \ref{ftrans}, we make two major improvements to make it applicable to our task. One improvement is that the denominator and numerator is the square root of the original one in Eq. \ref{gdn} which makes it is easy to implement the gradient propagation. Another improvement is to replace the response of filters $L_i$ with channel components of features $S_i$, which is proper for channel-wise feature disentanglement.

In order to guide the study of parameters from convolution layer, ${w}_{ji}$ and ${b}_{i}$, we introduce the spectral value difference orthogonality regularization (SVDO) from \cite{chen2019abd} as a loss constraint. As a novel method to reduce feature correlations, SVDO can be expressed as Eq. \ref{floss}.
\begin{equation}
\label{floss}
{Loss_F} = \left\| {{\lambda _1}\left( {F{F^T}} \right) - {\lambda _2}\left( {F{F^T}} \right)} \right\|_2^2
\end{equation}
where ${\lambda _1}\left( {F{F^T}} \right)$
and ${\lambda _2}\left( {F{F^T}} \right)$ denote the largest and smallest eigenvalues of ${F{F^T}}$, respectively. $F$ is feature maps and $T$ expresses the transposition.

\begin{figure}[htp]
	\centering
	\includegraphics[width=\linewidth]{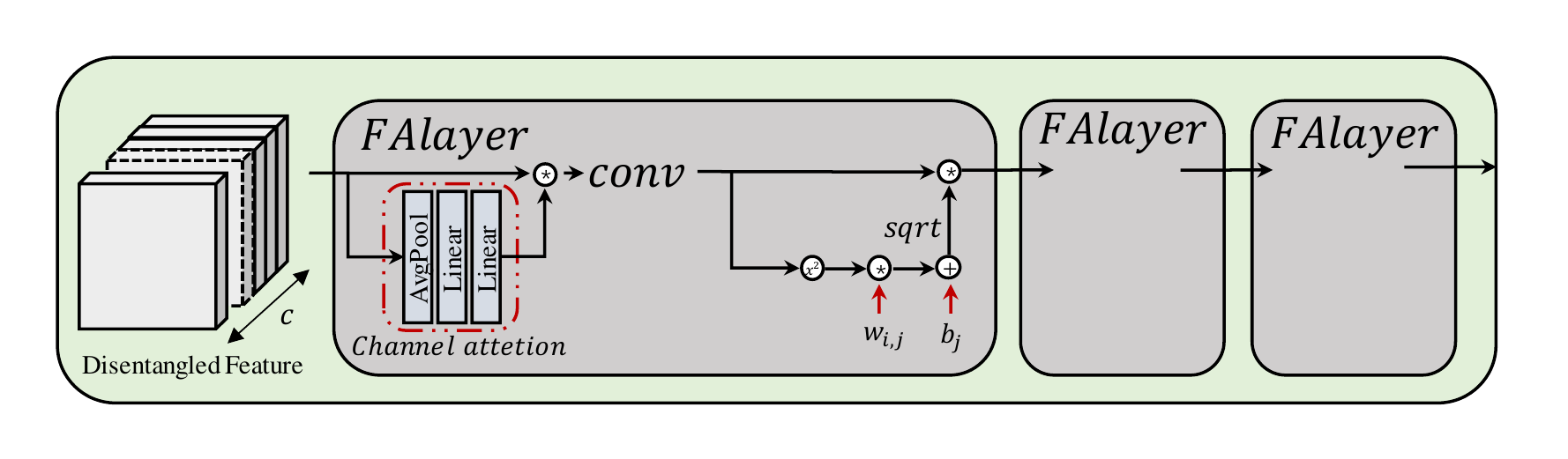}
	\caption{Feature Aggregation Module.}
	\label{fig:IFGmodel}
\end{figure}
According to above analysis, we design the corresponding feature disentanglement module (FDM). As seen in the Fig \ref{fig:FGmodel}, Each FDM have multiple FDlayers and each FDlayer is composed of one convolution layer and a revised gain-control-based normalization. 
Experiments demonstrate that three FDlayers for each FDM module are saturated to achieve the best performance as shown in Table. \ref{tab:ablation}.


\subsection{Feature aggregation Module}
For hybrid-distorted image restoration, we need to restore the image instead of only getting processed feature representation. To further filter out the distortion representation and maintain the raw image content details, we utilize channel-wise attention mechanism to adaptively select useful feature representation from processed disentangled feature as Eq. \ref{att}.

\begin{equation}
	\label{att}
	{F_{p\_i}} = PM({D_i}),\\
	{F_{p\_i}} = CA({F_{p\_i}}),
\end{equation}
where PM represents the process module and CA presents channel attention. $D_i$ is the $ith$ channel of disentangled feature. $F_{p\_i}$ represents the $i$th channel of output feature.
To construct the image, we get the inversion formula  from  Eq. \ref{ftrans} as: 
\begin{equation}
{{F}_{c\_i}}={{F}_{p\_i}}\sqrt{\sum\nolimits_{j}{{{w}_{ji}}F_{p\_j}^{2}+{{b}_{i}}}},
\label{IDG}
\end{equation}
where ${F}_{c\_i}$ represents the output feature corresponding to the distribution of clean image. 
With this module, the processed image information could be aggregated to original feature space, which is proper for reconstructing restored image.
Feature aggregation module (FAM) is designed as shown in Fig. \ref{fig:IFGmodel}

\subsection{Auxiliary Module}
In the processes of feature disentanglement, the mutual information of different channels of features is reduced, which will result in some loss of image information. In order to make up for the weakness of the feature disentanglement branch, we used the existing ResBlock to enhance the transmission of image information in parallel. 
\begin{figure}[htp]
	\centering
	\includegraphics[width=\linewidth]{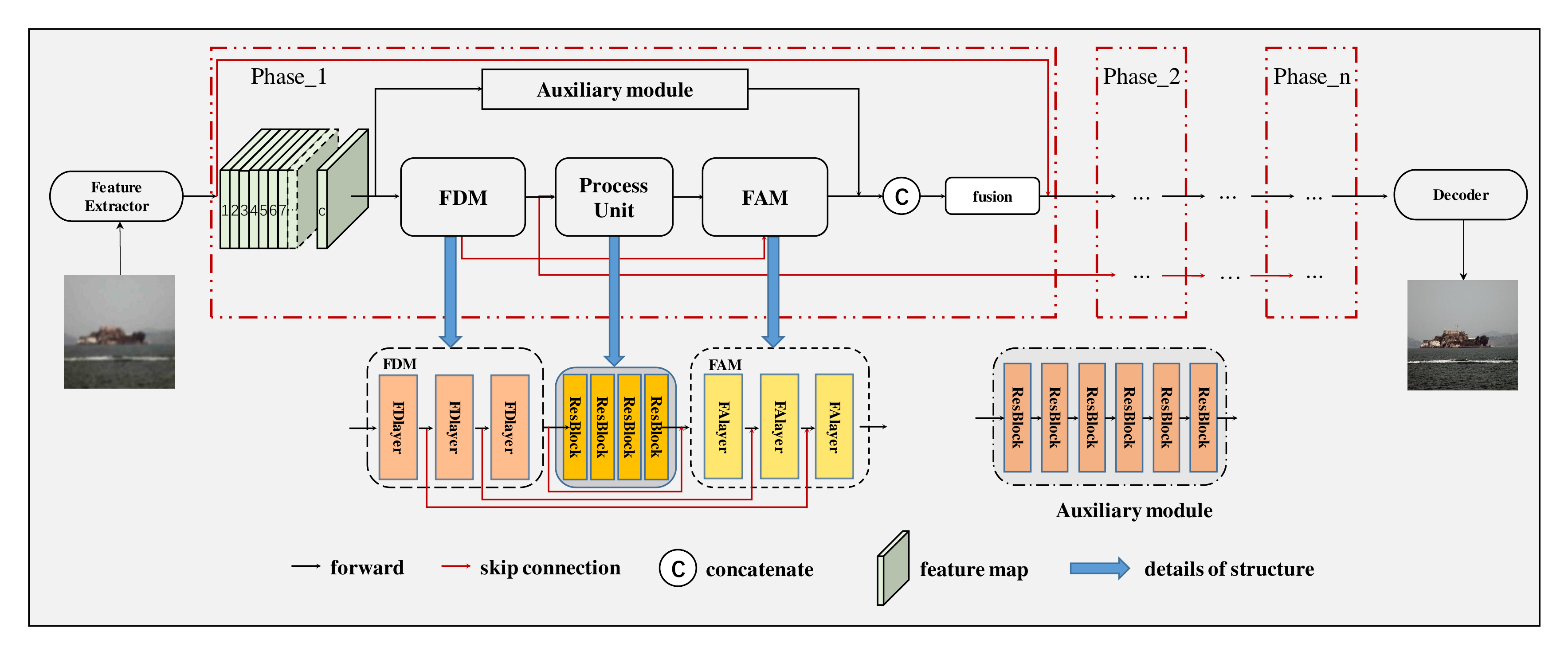}
	\caption{Architecture of our proposed FDR-Net. It consists of multiple phases and each phase consists of feature disentangle module (FDM) and feature aggregation module (FAM).}
	\label{fig:framework}
\end{figure}
\subsection{Overview of Whole Framework}
To make full use of feature information of different levels, we use multi-phases structure to deal with hybrid distortion as Fig. \ref{fig:framework}. The whole structure is composed of several same phases. Every phase is composed by two branches, where the lower branch is used for distortion-aware feature disentangle and feature processing, the upper branch is used as auxiliary branch which is used to make up for information lost and enhance the removal of hybird distortion. The details of lower branch are shown in Fig. \ref{fig:framework}. In order to ensure the matching between FDlayer and FAlayer, we use skip connection to connect the output of FDlayer and the input of FAlayer. To fuse the information of outputs from two branches, we first concatenate the two outputs and then use channel attention and convolution to fuse the outputs to a feature. Furthermore, we use dual residual to relate the different phases to enhance the mutual utilization of information between different phases.  
\subsection{Loss Function.}
In this paper, the loss function is composed of L1 loss and feature orthogonality loss with the weight $\beta$ 0.00001 as Eq. 5. 
The loss function can be expressed as:
\begin{equation}
Loss = L1loss + \beta los{s_F}
\end{equation}
\section{Experiments}
In this section, we first conduct extensive experiments on DIV2K dataset of hybrid distortions \cite{yu2018crafting} and results show that our FDR-Net achieves superior performance over the state-of-the-art methods. Moreover, in order to increase the difficulty of hybrid-distorted image restoration, we construct a more complex hybrid distortion dataset DID-HY based on DID-MDN dataset \cite{zhang2018density} and test our approach on it. Then we validate the robustness of our FDR-Net by conducting extensive experiments on single distortion. 
To make our network more explanatory, the correlation maps of features are visualized. Finally, we conduct a series of ablation study to verify the effectiveness of our modules in FDR-Net.  
\begin{table}[htp]
	\centering
	\caption{Quantitative results on the DIV2K dataset comparison to the state-of-the-art methods in terms of PSNR and SSIM. OP-Attention represents the operation-wise attention network in \cite{suganuma2019attention}. The SSIM is computed in the same way with OP-Attention.}
	\label{tab:hybrid}
	\begin{tabular}{c|cc|cc|cc}
		\hline
		Test set                          & \multicolumn{2}{c|}{Mild(unseen)}                                     & \multicolumn{2}{c|}{Moderate}                                         & \multicolumn{2}{c}{Severe(unseen)}                                   \\ \hline
		Metric                           & PSNR                        & SSIM                          & PSNR                        & SSIM                          & PSNR                         & SSIM                         \\ \hline
		DnCNN \cite{zhang2017beyond}                           & 27.51                                 & 0.7315                                 & 26.50                                 & 0.6650                                 & 25.26                                 & 0.5974                                 \\
		\multicolumn{1}{l|}{RL-Restore\cite{yu2018crafting}}   & 28.04                                 & 0.7313                                 & 26.45                                 & 0.6557                                 & 25.20                                 & 0.5915                                 \\
		\multicolumn{1}{l|}{OP-Attention\cite{suganuma2019attention}} & 28.33                                 & 0.7455                                 & 27.07                                 & 0.6787                                 & 25.88                                 & 0.6167                                 \\
		Ours                           & {\textbf{28.66}} & { \textbf{0.7568}} & { \textbf{27.24}} & {\textbf{0.6846}} & { \textbf{26.05}} & { \textbf{0.6218}} \\ \hline
	\end{tabular}
\end{table}

\subsection{Dataset for hybrid-distorted image restoration}

\textbf{DIV2K dataset.} The standard dataset on hybrid distortion are produced in \cite{yu2018crafting} based on DIV2K dataset. The first 750 images are used for training and remaining 50 images are used for testing. They first process the dataset by adding multiple types of distortions, including Gaussian noise, Gaussian blur, and JPEG compression artifacts. The standard deviations of Gaussian blur and Gaussian noise are randomly chosen from the range of [0, 10] and [0, 50]. The quality of JPEG compression is chosen from [10, 100]. According to the degradation level, the dataset is divided into three parts, respectively are mild, moderate, and severe which represent that the degradation level from low to high.
Then they crop the patches of size $63\times63$ from these processed images as training set and testing set. The training set consists of 249344 patches of moderate class and testing sets  consists of 3584 patches of three classes. 

\begin{figure}[htp]
	\centering
	\includegraphics[width=0.9\linewidth]{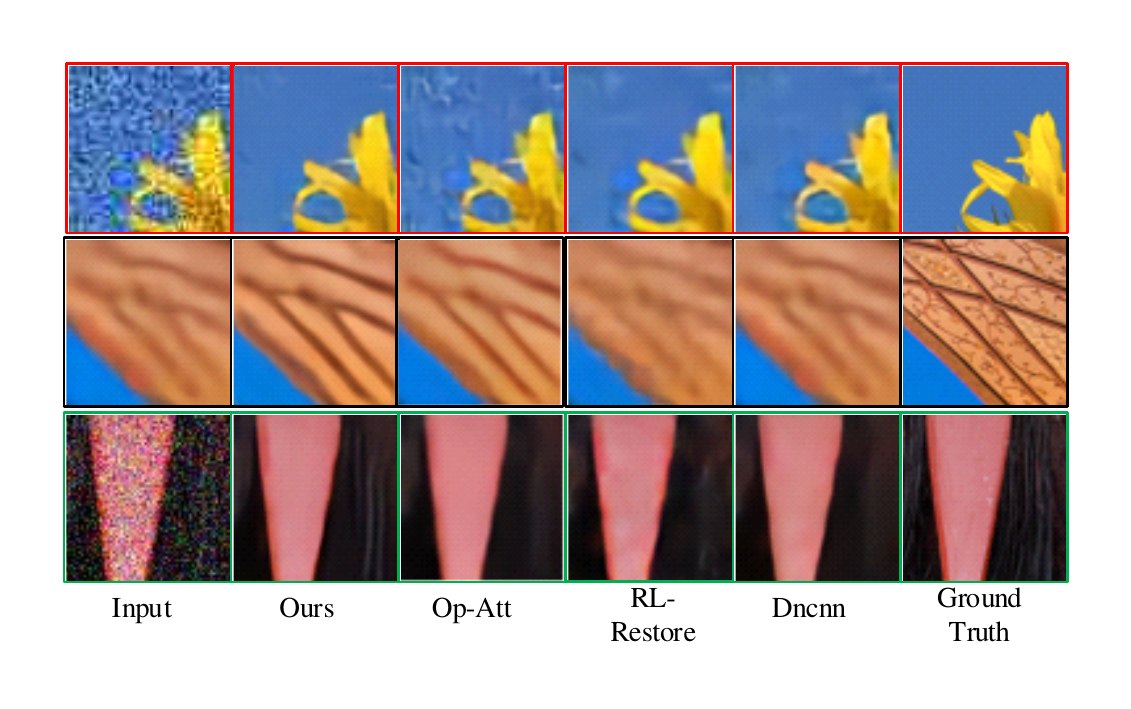}
	\caption{The performance comparison of our algorithm with the stat-of-the-art methods performed on DIV2K dataset. The three groups are corresponding to severe, moderate and mild  from top to bottom.}
	\label{fig:results}
\end{figure}
\begin{table}[h]
	\centering
	\caption{Quantization results on DID-HY dataset compared with the state-of-the-art methods. The SSIM is computed in the same way with OP-Attention.}
	\label{mix_2}
	\small
	\begin{tabular}{c|c|c|c}
		\hline
		Method & Dncnn\cite{zhang2017beyond} & OP-Attention \cite{suganuma2019attention}& ours \\ \hline
		PSNR   &   22.1315             & 23.1768               & \textbf{24.9700}      \\ \hline
		SSIM  &   0.5953             & 0.6269                & \textbf{0.6799}       \\ \hline
	\end{tabular}
\end{table}
\subsubsection{DID-HY dataset.} Since the resolution of test image in DIV2K is too low, which is not practical in real world, we design a more complicated dataset which contains four kinds of distortions and higher resolution. The new dataset is built by adding Gaussian blur, Gaussian noise and JPEG compression artifacts based on DID-MDN dataset \cite{zhang2018density}. The training set contains $12,000$ distortion/clean image pairs, which have resolutions of $512\times512$. And the testing set contains $1,200$ distorted/clean image pairs.
\subsection{Datasets for single distortion restoration}
                       
\subsubsection{Gopro dataset for Deblurring.}  
As the standard dataset for image deblurring, GOPRO datast is produced by \cite{kupyn2018deblurgan}, which contains 3214 blurry/clean image pairs.  The training set and testing set consists of 2103 pairs and 1111 pairs respectively.
\subsubsection{DID-MDN Dataset for Deraining.}
The DID-MDN dataset is produced by \cite{zhang2018density}.  Its training dataset consists of 12000 rain/clean image pairs which are classified to three level (light, medium, and heavy) based on its rain density. There are 4000 image pairs per level in the dataset. The testing dataset consists of 1200 images, which contains rain streaks with different orientations and scales.

\subsection{Implementation Details}
The implement of the proposed framework is based on PyTorch framework with one NVIDIA 1080Ti GPU. In the process of training, we adopt Adam optimizer \cite{kingma2014adam} with a learning rate of 0.0001. The learning rate will decay by a factor valued 0.8 every 9000 iterations and the number of mini-batches is 28. It takes about 24 hours to train the network and get the best results.

\begin{figure}[htp]
	\centering
	\includegraphics[width=0.95\linewidth]{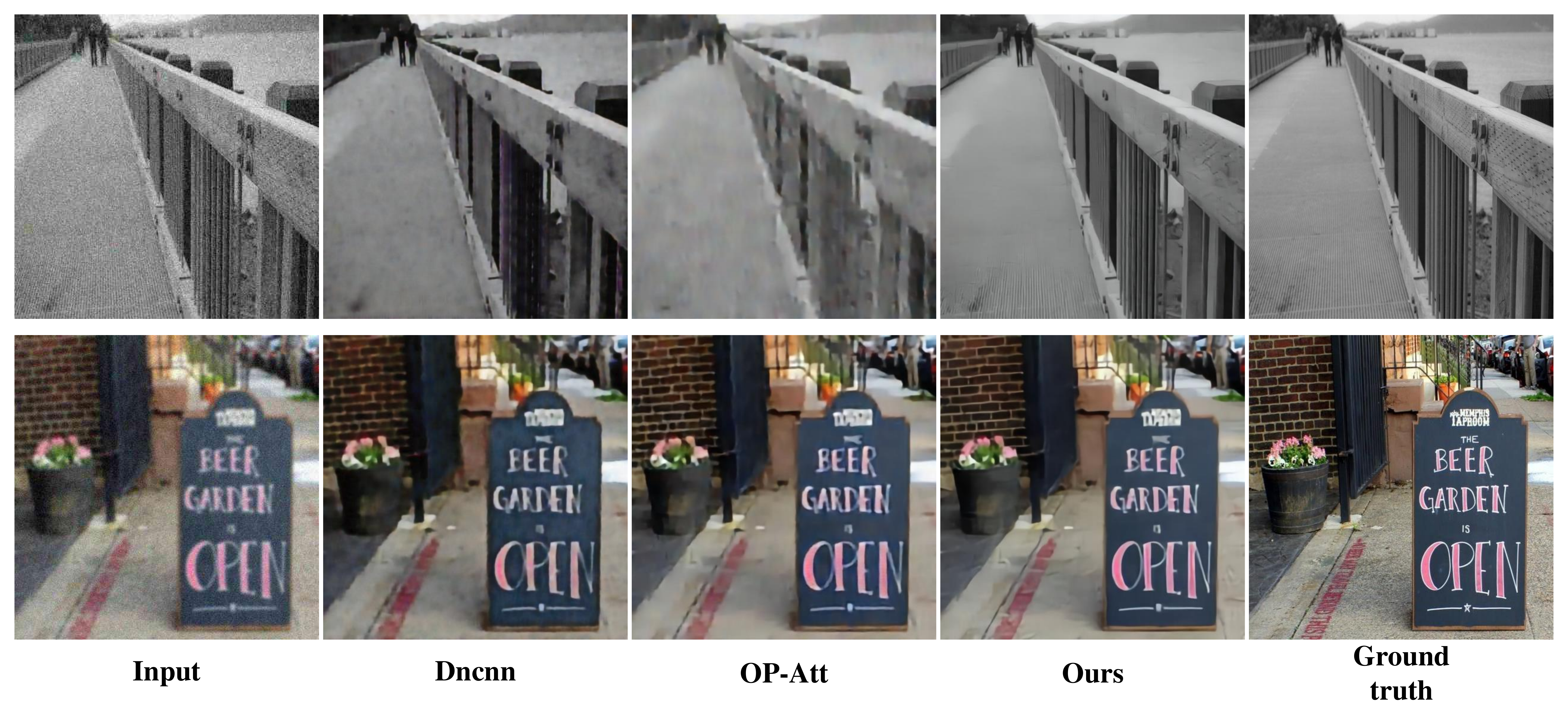}
	\caption{Examples of hybrid distortion removal on DID-HY dataset.}
	\label{fig:results_mix2}
\end{figure}

\begin{table}[h]
	\centering
	\caption{Quantitative results on Gopro dataset compared with the state-of-the-art methods for image deblur. The SSIM is computed in the same way with SRN \cite{tao2018scale}.}
	\label{tab:blur}
	\begin{tabular}{c|c|c|c|c|c}
		\hline
		Method & DeepDeblur\cite{wang2017deepdeblur} & DeblurGan \cite{kupyn2018deblurgan} & SRN \cite{tao2018scale}& Inception\_ResNet\cite{kupyn2019deblurgan} &  ours \\ \hline
		PSNR  & 29.23               & 28.70              & 30.10        & 29.55                          & \textbf{30.55}    \\ \hline
		SSIM  & 0.916               & 0.927              & 0.932        & 0.934                         & \textbf{0.938}    \\ \hline
	\end{tabular}
\end{table}


\begin{figure}[h]
	\centering
	\includegraphics[width=\linewidth]{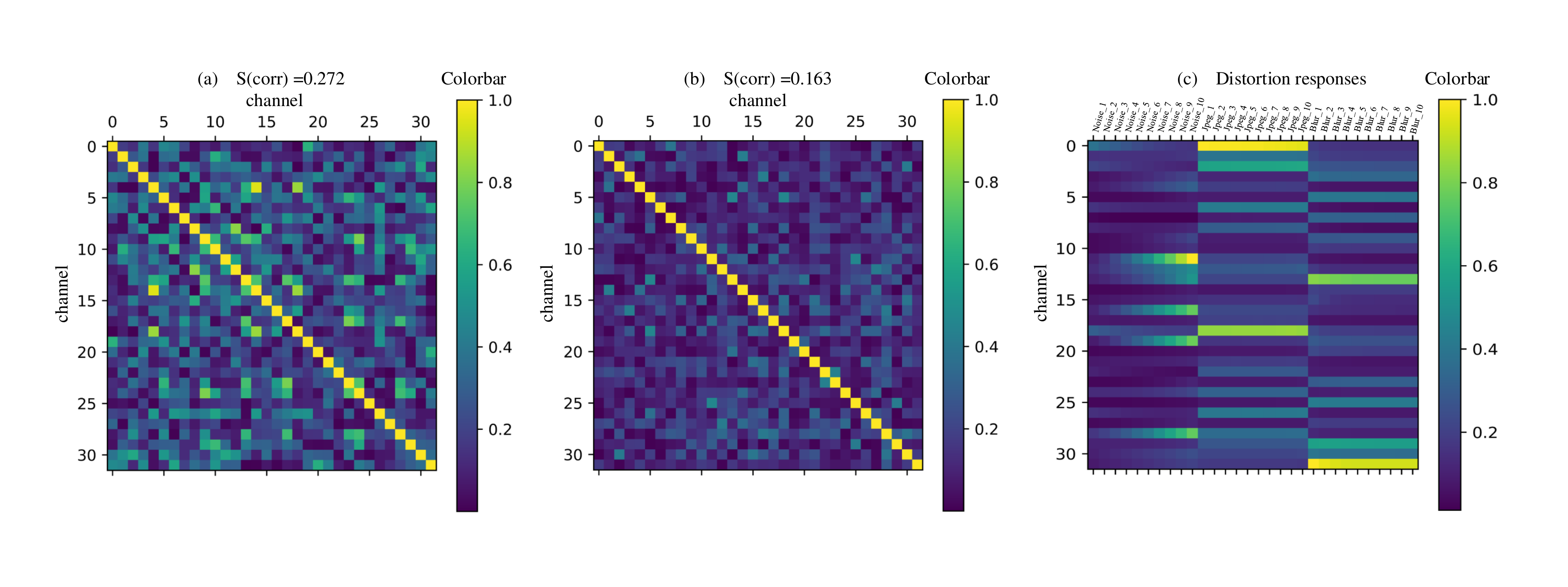}
	\caption{(a)Visualization of correlation matrix between channels from the feature before FDM. (b) Visualization of correlation matrix between channels from feature after FDM. (c) Channels responses corresponding to different distortion types after FDM. As shown in (a)(b), FDM reduces the channel-wise correlations by disentangling the feature. From (c), the different distortions are distributed in different channels regardless of the levels of distortion by feature disentanglement. }
	\label{fig:vis}
\end{figure}.

\subsection{Comparison with State-of-the-Arts}
In this section, we compare the performance of our FDR-Net on the DIV2K dataset with  state-of-the-art methods, including RL-Restore \cite{yu2018crafting} and operation-wise attention network \cite{suganuma2019attention}. Since DnCNN \cite{zhang2017beyond} was used as baseline in previous work, we compared with it too. Table. \ref{tab:hybrid} have demonstrated the superiority of our method. As shown in Table. \ref{tab:hybrid}, our algorithm outperforms previous work in mild, moderate and severe level of hybrid distortion just training with moderate level dataset. And the  comparison of subjective quality is shown in the Fig. \ref{fig:results}.

Since the size of the image in above dataset is just 63$\times$63 and contains fewer distortion types, which is not practical in real world, we synthesize a more complex hybrid distortion dataset named DID-HY based on DID-MDN dataset. This dataset contains four kinds of distortions including blur, noise, compression artifacts and rain, in which  the resolution of image is 512$\times$512. Then we retrained the DnCNN\cite{zhang2017beyond}, operation-wise attention network\cite{suganuma2019attention} and our FDRNet. The results in Table. \ref{mix_2} have demonstrated that the Dncnn\cite{zhang2017beyond}, operation-wise attention network \cite{suganuma2019attention} can't work well for more complicated hybrid distortion with higher resolution. However, our FDRNet can obtain the best results in objective quality as shown in Table. \ref{mix_2} and subjective quality as shown in Fig. \ref{fig:results_mix2}.

\subsection{Interpretative Experiment}
To visualize the channel-wise feature disentanglement, we study the correlation matrix between the channel of features before and after FDM. The feature before FDM reveal high correlations(0.272 in average) as shown in Fig. \ref{fig:vis} (a). By disentangling the feature with our FDM, the feature correlations are suppressed as Fig. \ref{fig:vis} (b) (0.164 in average). Moreover, we visualize the channel response of different distortion types after FDM in Fig.\ref{fig:vis} (c). We take three kinds of distortions including blur, noise, and jpeg artifacts. And Each kind of distortion contains ten kinds of distortion levels. The higher level indicates the larger distortion. As shown in Fig.\ref{fig:vis}(c), different kinds of distortions have different channel responses regardless of the distortion levels, which means that the distortions have been divided across different channels by feature disentanglement. Besides, the different distortion levels only bring the changes of response strength.
\subsection{Experiments on Single Distortion}

In order to verify the robustness of our method, we separately apply our FDR-Net on image deblurring  and  image deraining tasks and achieved the stat-of-the-art performance. 
\begin{figure}[htp]
	\centering
	\includegraphics[width=0.95\linewidth]{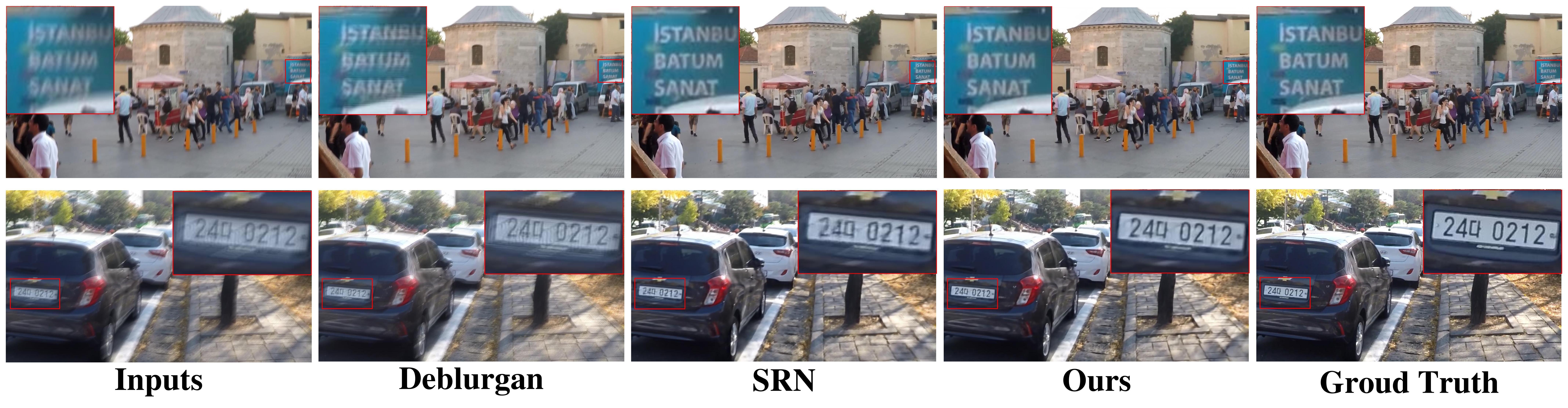}
	\caption{Visual comparisons for Gopro-test dataset between our FDR-Net and the state-of-art methods for image deblurring.}
	\label{fig:deblur}
\end{figure}

\begin{table}[htp]
	\centering
	\caption{Quantitative results on DID-MDN dataset compared with the state-of-the-art methods for image derain. The SSIM is computed in the same way with SPANET \cite{wang2019spatial}.}
	\setlength{\tabcolsep}{0.25mm}{
		\begin{tabular}{c|c|c|c|c}
			\hline
			Method & DID-MDN \cite{zhang2018density} & RESCAN\cite{li2018recurrent} & SPANET \cite{wang2019spatial} & ours \\ \hline
			PSNR   & 27.95   & 29.95  & 30.05   & \textbf{32.96}        \\ \hline
			SSIM   & 0.909   & 0.884  & 0.934   & \textbf{0.951}             \\ \hline
		\end{tabular}
	}
	\label{tab:derain}
\end{table}

\textbf{Comparison to image deblurring.} On the task of image deblur, we compared our FDR-Net with DeepDeblur\cite{wang2017deepdeblur}, Deblurgan \cite{kupyn2018deblurgan}, SRN \cite{tao2018scale} and Deblurganv2 \cite{kupyn2019deblurgan} with Inception ResNet respectively. The table. 3 have demonstrate that our FDR-Net surpasses previous works for image deblurring.  As seen in the Fig. \ref{fig:deblur}, our algorithm is better than other state-of-the-art methods in generating rich texture.

\begin{figure}[htp]
	\centering
	\includegraphics[width=0.95\linewidth]{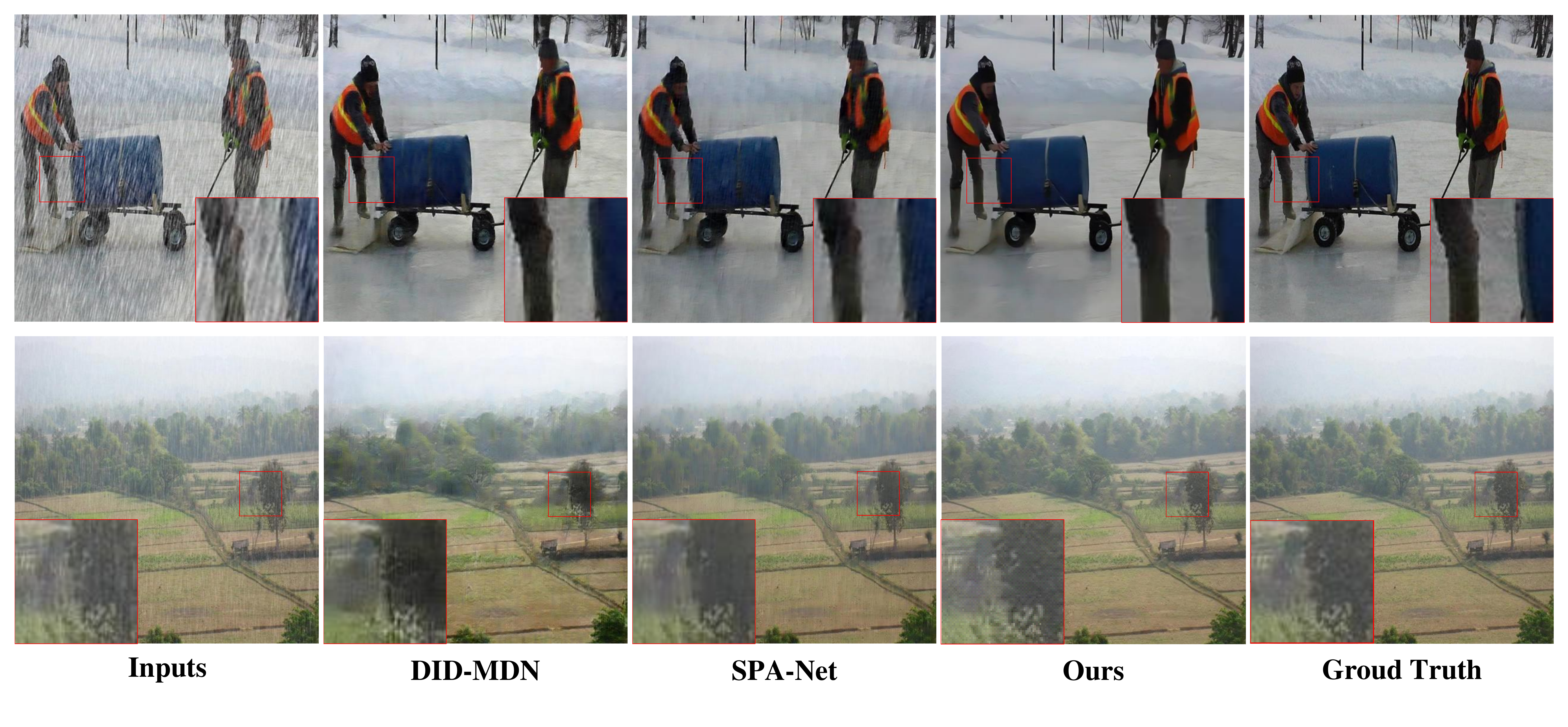}
	\caption{Visual comparisons for DID-MDN dataset between our FDR-Net and the state-of-art methods for image deraining.}
	\label{fig:derain}
\end{figure}

\textbf{Comparison to image deraining.} In the field of image derain, we compared our FDR-Net with DID-MDN\cite{zhang2018density}, RESCAN\cite{li2018recurrent} and SPANET\cite{wang2019spatial}. The quantization quality as shown in Tab. \ref{tab:derain} have demonstrated the  
superiority of our method for image deraining task. The subjective quality is compared with the state-of-the-art methods in Fig. \ref{fig:derain}.
\subsection{Ablation Studies}
\subsubsection{Effect of FDM.}
Distortion-aware feature disentangle method is a key factor to the success of FDR-Net for hybrid-distorted image restoration. To verify its importances and effectiveness, we substitute the FDlayer and FAlayer with a series of ResBlocks\cite{he2016deep} as our baseline. Then we retrained the baseline network with DID-HY dataset. The parameters of two networks are almost the same. However, the performance of our FDR-Net is far higher than the baseline without FDlayer and FAlayer. The quantitative results are shown in Table. \ref{tab:no_dfd}.

\begin{table}[!htp]
	\centering
	\caption{Ablation study on different combinations of FDM and Auxiliary module. Tested on DID-HY dataset.}
	\label{tab:no_dfd}
	\setlength{\tabcolsep}{4mm}{
		\begin{tabular}{cc|ccc}
			\hline
			FDM. & Aux. & Params & PSNR & SSIM \\ \hline
			$\checkmark$& $\checkmark$  &  1.66M & 24.97 & 0.680 \\
			$\times$& $\checkmark$ & 1.62M & 23.91 & 0.646\\
			$\checkmark$&$\times$  & 1.21M & 24.27 & 0.653 \\ \hline
		\end{tabular}
	}
\end{table}

\subsubsection{Effect of multiple FDlayers and multiple channels.}
To investigate the effect of multiple FDlayers in FDM, we set the number of FDlayers as 1 to 5 respectively. As shown in Table. \ref{tab:ablation} , three FDlayers are saturated to achieve the best performance. Then we set the number of FDlayers as 3 and set the number of channels as 16, 32, 48 and 64 respectively.  As shown in Table. \ref{tab:ablation}, increasing the number of channels could increase the ability to represent the distortions for Network, which could bring the improvement of performance and 32 is saturated to achieve the best performance.
\begin{table}[htp]
	\centering
	\caption{Comparisons between different number of FDlayers in each FDM and comparisons between different number of channels with 3 FDlayers.}
	\begin{tabular}{c|c|c|c|c|c|c|c|c|c|c}
		\hline
		DFlayers & 1    & 2 & 3 & 4 & 5 & Channels & 16 & 32 & 48 & 64\\ \hline
		PSNR     & 24.63 &  24.70 & 24.97  & 24.98  &  24.99  & PSNR     & 24.47   &  24.97  & 24.98  & 25.00\\ \hline
		SSIM     & 0.666  & 0.668  & 0.680  & 0.680  & 0.681  & SSIM     & 0.659   & 0.680   & 0.682  & 0.683\\ \hline
	\end{tabular}
	\label{tab:ablation}
\end{table}
\subsubsection{Effect of Auxiliary Module.}
As discussed in Section 3, auxiliary module would make up for the weakness of the distortion-aware feature disentangle branch. We demonstrate the effectiveness of auxiliary module by removing it. The performance reduction is shown as Table \ref{tab:no_dfd}.
\subsubsection{Effect of Multi-phases.}
In order to demonstrate the effect of  multi-phases, we set the number of phases from 2 to 7. As shown as Table. \ref{tab:multi-steps}, more phases will bring better performance. However, the performance is near saturation when the number of phases is 6. 

\begin{table}[!htp]
\centering
\label{tab:multi-steps}
\caption{A comparison of our methods with different phases.}
\begin{tabular}{c|c|c|c|c|c|c}
\hline
Phases &    2&3 & 4 & 5 & 6 &7\\ \hline
params &    0.59M & 0.85M      & 1.12M       & 1.39M      & 1.66M & 1.93      \\ \hline
PSNR   &    24.20 & 24.42     & 24.60      & 24.71      & 24.97 & 24.98   \\ \hline
SSIM   &    0.651 & 0.658    & 0.665      & 0.668      & 0.680 & 0.681  \\ \hline
\end{tabular}
\label{tab:multi-steps}
\end{table}

\section{Conclusion}
In this paper, we implement the channel-wise feature disentanglement by FDM to reduce the interference between hybrid distortions, which achieves the best performance for hybrid-distorted image restoration. Furthermore, we also validate the effectiveness of our modules with the visualizations of correlation maps and channel responses for different distortions. Extensive experiments demonstrate that our FDR-Net has stronger robustness for single distortion removal.

\section{Acknowledgement}
This work was supported in part by NSFC under Grant
U1908209, 61632001 and the National Key Research and
Development Program of China 2018AAA0101400.
\clearpage
%
%
\bibliographystyle{splncs04}
\bibliography{egbib}
\section{Appendix}

In sec. 7.1, we show the interference between hybrid distortions by a series of examples.
In sec. 7.2, we show the interference between hybrid distortions at feature level. 
\subsection{Examples of interference in hybrid distortions}
In this section, we show the interference between hybrid distortions by a series examples. 
\begin{figure}[htp]
	\centering
	\includegraphics[width=\linewidth]{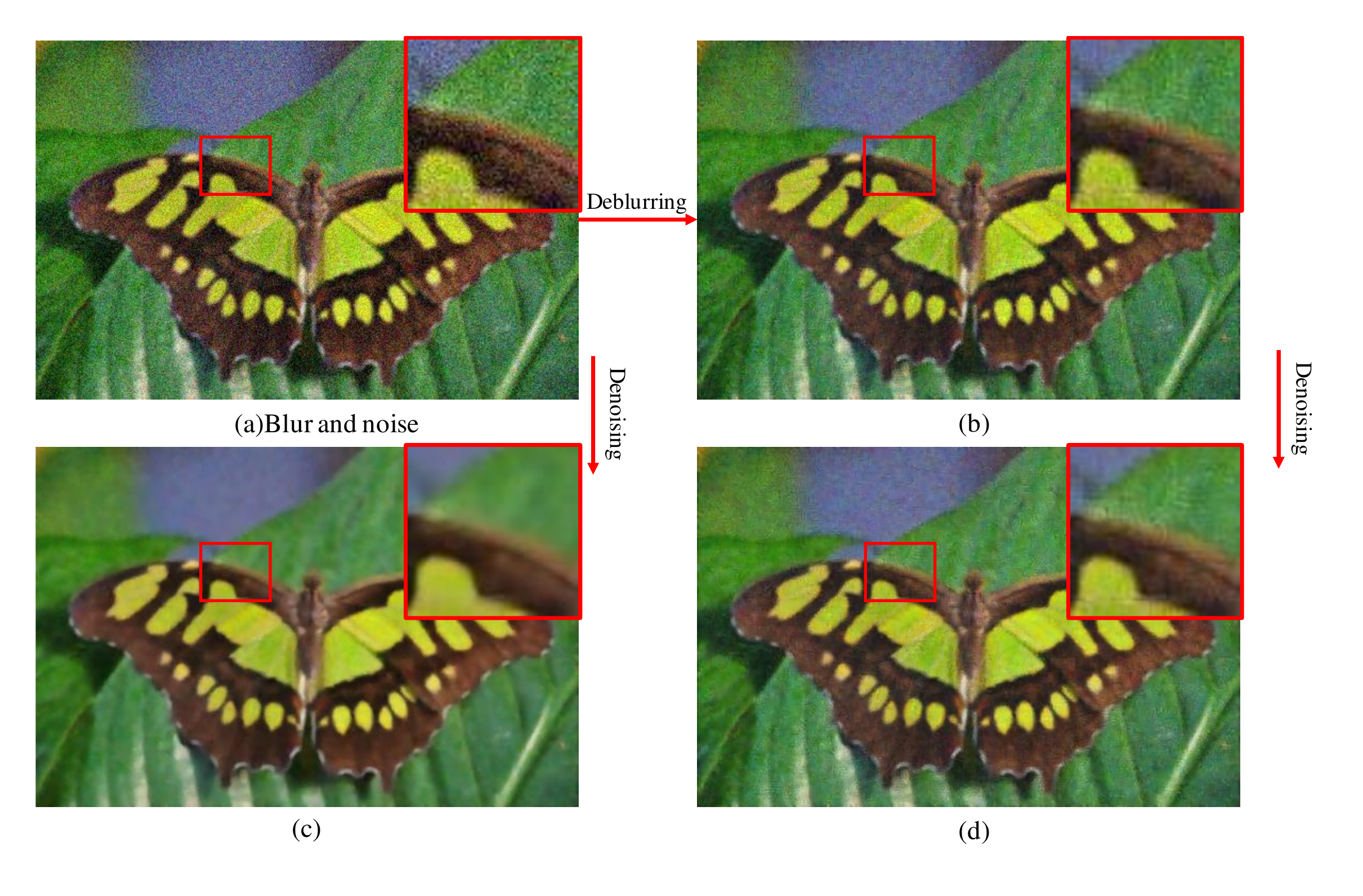}
	\caption{Examples of interference between blur and noise. (a) Distorted image with blur and noise.   (b) Image obtained by deblurring (a).   (c) Image obtained by denoising (a). (d) Image obtained by denoising (b).}
	\label{fig:example_a}
\end{figure}
\subsubsection{The interference between blur and noise}
As shown in Fig. \ref{fig:example_a}, the Fig. \ref{fig:example_a}(a) contains two kinds of distortions, including noise and JPEG artifacts . If we denoise the Fig. \ref{fig:example_a}(a) directly, the noise can be removed well as Fig. \ref{fig:example_a}(c). However, if we first make deblurring for Fig. \ref{fig:example_a}(a) as Fig. \ref{fig:example_a}(b), the distribution of noise would be changed, and we cannot remove the noise well as Fig. \ref{fig:example_a}(d). According to above experiments, we can find that the noise could be interfered when we process the blur, which makes it difficult to process the noise. 
\begin{figure}[h]
	\centering
	\includegraphics[width=\linewidth]{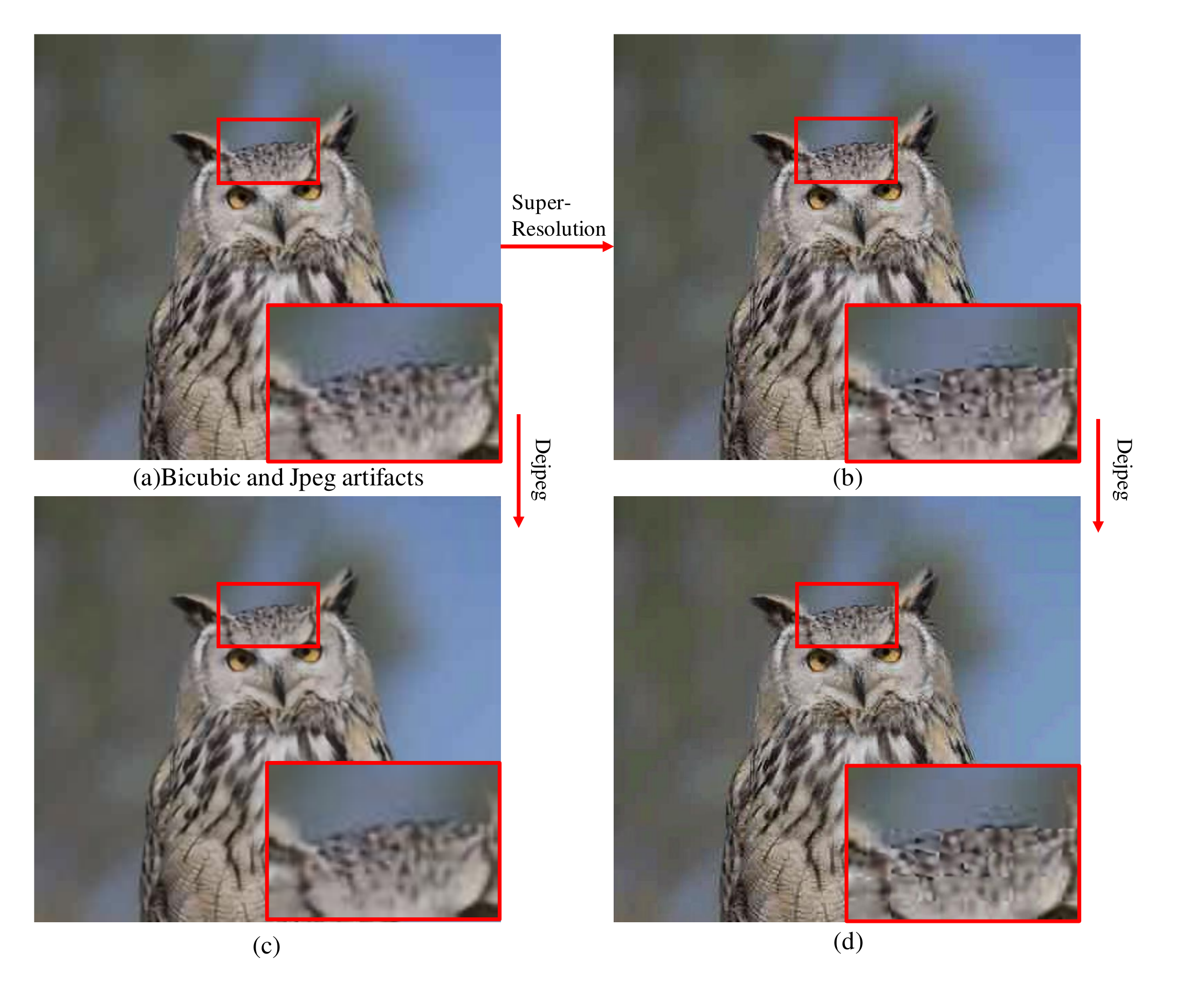}
	\caption{Examples of interference between low-resolution and JPEG artifacts. (a) Distorted image with low-resolution and JPEG artifacts. (b) Image obtained by making super-resolution for (a). (c) Image obtained by removing JPEG artifacts for (a). (d) Image obtained by removing JPEG artifacts for (b).}
	\label{fig:example_b}
\end{figure}

\subsubsection{The interference between low-resolution and JPEG artifacts}
As shown in Fig. \ref{fig:example_b}, Fig. \ref{fig:example_b}(a) contains two kinds of distortions, including low-resolution and JPEG artifacts. If we remove the JPEG artifacts for Fig. \ref{fig:example_b}(a), the JPEG artifacts could be removed well. However, if we first make super-resolution for Fig. \ref{fig:example_b}(a) as Fig. \ref{fig:example_b}(b). We can find that the JPEG artifacts are amplified as shown in  \ref{fig:example_b}(b) when we make super-resolution, which causes that the JPEG artifacts cannot be removed well as Fig. \ref{fig:example_b}(d). Therefore, different distortions could be interfered with each other in hybrid distortions, which increase the difficulty of hybrid-distorted image restoration.

\subsection{Visualization of interference in different distortions.}
\label{sec2}
In this section, we visualize the interference between hybrid distortions at feature level. We first train a regular CNN network DnCNN \cite{zhang2017beyond} with the datasets of hybrid distortions in \cite{yu2018crafting}. Then we visualize the channel responses of different distortions at feature level. We take three kinds of distortions including blur, noise, and JPEG artifacts. And each kind of distortions contains ten kinds of distortion levels. The higher level indicates the larger distortion. As shown in Fig. \ref{fig:distortion_example}, different kinds of distortion types are almost distributed in the same channels of features which are marked with red triangle. We can find that the features of different distortions are entangled, which means the interference between hybrid distortions at feature level. Therefore, we proposed to learn disentangled features for hybrid-distorted image restoration. The effectiveness of our approach could be seen in Fig. 8(c) in our paper. 

\begin{figure}[h]
	\centering
	\includegraphics[width=\linewidth]{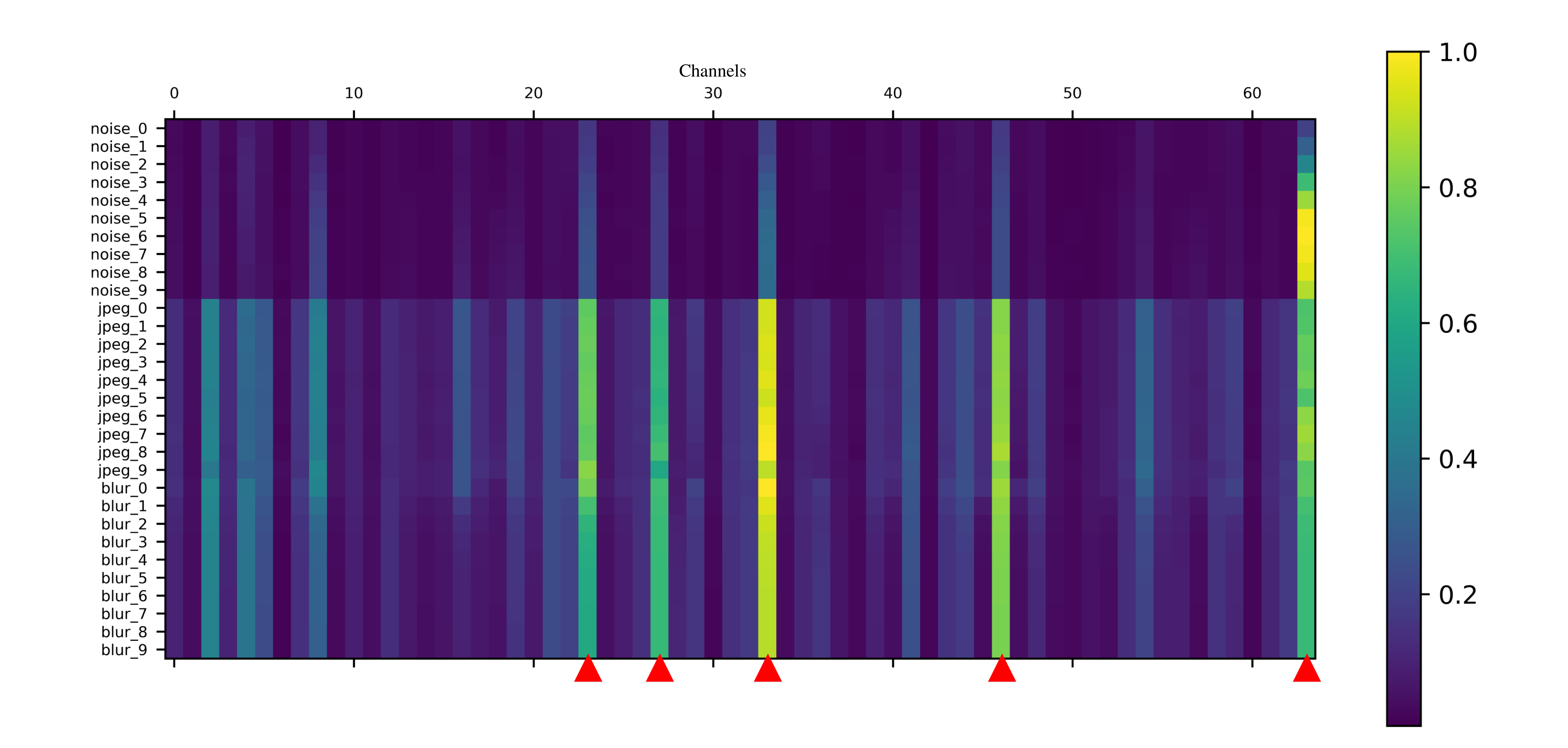}
	\caption{Channel responses corresponding to different distortion types in features.}
	\label{fig:distortion_example}
\end{figure}

\end{document}